\documentclass[11pt,a4paper]{article}
\usepackage[hyperref]{emnlp2018}
\usepackage{times}
\usepackage{latexsym}
\usepackage{microtype}
\usepackage{amsmath}
\usepackage{amsfonts}
\usepackage{algorithm}
\usepackage[small]{caption}
\usepackage[noend]{algpseudocode}
\usepackage{booktabs}
\usepackage{adjustbox}
\usepackage{url}
\usepackage{supertabular}
\usepackage{xcolor}
\definecolor{brickred}{rgb}{0.8, 0.25, 0.33}
\definecolor{darkmidnightblue}{rgb}{0.0, 0.2, 0.4}

\usepackage{cleveref}
\crefname{section}{\S}{\S\S}
\Crefname{section}{\S}{\S\S}
\crefname{table}{Tab.}{}
\crefname{figure}{Fig.}{}
\crefname{algorithm}{Alg.}{}
\crefname{equation}{Eq.}{}
\crefname{appendix}{App.}{}
\crefformat{section}{\S#2#1#3}  

\usepackage[disable]{todonotes}

\newcommand{\note}[4][]{\todo[author=#2,color=#3,size=\scriptsize,fancyline,caption={},#1]{#4}} 
\newcommand{\katha}[2][]{\note[#1]{Katha}{yellow!40}{#2}}
\newcommand{\ryan}[2][]{\note[#1]{Ryan}{violet!40}{#2}}
\newcommand{\Ryan}[2][]{\ryan[inline,#1]{#2}}   
\newcommand{\lawrence}[2][]{\note[#1]{Lawrence}{cyan!40}{#2}}
\newcommand{\arya}[2][]{\note[#1]{Arya}{lime!40}{#2}}
\newcommand{\Arya}[2][]{\arya[inline,#1]{#2}}   
\newcommand{\word}[1]{\textit{#1}}

\newcommand{\set}[1]{\left\{ #1 \right\}}

\aclfinalcopy 

\title{Grammatical Gender, Neo-Whorfianism, and Word Embeddings: A Data-Driven Approach to Linguistic Relativity}

\author{  
  Katharina Kann\\ 
  New York University, USA\\
  \texttt{kann@nyu.edu} 
}

\date{}
\begin{document}
\maketitle
\begin{abstract}
The relation between language and thought has occupied linguists for
at least a century. Neo-Whorfianism, a weak version of the controversial Sapir--Whorf
hypothesis, holds that our thoughts are subtly influenced by 
the grammatical structures of our native language.
One area of investigation in
this vein focuses on how the grammatical gender of nouns
affects the way we perceive the corresponding objects.
For instance, does the fact that \word{key} is masculine in German
(\word{der Schl{\"u}ssel}), but feminine in Spanish (\word{la llave})
change the speakers' views of those objects?  Psycholinguistic evidence
presented by \newcite[\S 4]{boroditsky2003sex} suggested the answer might be yes: When asked to produce 
adjectives that best described a \word{key}, 
German and Spanish speakers named more stereotypically masculine and feminine ones, respectively.
However, recent attempts to replicate those experiments have
failed \cite{mickan2014key}. 
In this work,  
we offer a 
computational analogue of \newcite[\S 4]{boroditsky2003sex}'s experimental
design on 9 languages, 
finding evidence \emph{against} neo-Whorfianism.
\end{abstract}

\section{Introduction}\label{sec:introduction}
During his tenure as a graduate student, 20$^\text{th}$-century
American linguist Benjamin Whorf conducted field work on Hopi, an
Uto-Aztecan language spoken in Southern Arizona. To his surprise, he
found that Hopi does not mark the tense of a verb in the way
many Western European languages do
\cite{whorf1956hopi}.\footnote{Whorf's claim has subsequently been
  challenged. Later analyses of Hopi grammar suggest that
  the language marks two tenses: future and non-future
  \cite{malotki1983hopi}.} Thus, according to Whorf, a Hopi speaker
must infer whether an action takes place in the past, present or
future only from the sentential context in which the verb
occurs. This finding inspired Whorf to start questioning whether
language influences thought, a position that has come to be known as linguistic relativity
\cite{whorf2012language}.  Ultimately, Whorf went on to hypothesize
that the Hopi perceive time differently as a result of their
language's grammar, kicking off a more encompassing debate on the relation of
language and thought and engendering one of the larger controversies in
linguistics to date \cite{deutscher2010through}. 

\begin{figure*}
\centering
  \includegraphics[width=\textwidth]{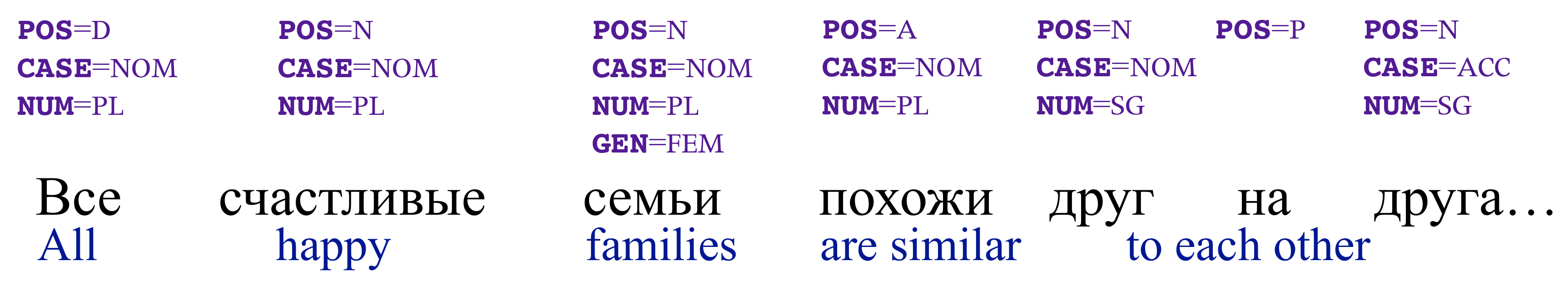}
  \caption{A Russian sentence showing the original forms and their decomposition into tag--lemma pairs.}
  \label{fig:sentence}
\end{figure*}

While influential, the strong version of what has come to be known as
the Sapir--Whorf\arya{Everywhere, Sapir--Whorf should be written with an en-dash. I've changed this.}\footnote{The hypothesis is named after both Benjamin Whorf and his
  Ph.D. advisor Edward Sapir.}  hypothesis has been disproven. For instance, even though
languages differ in the basic color terms they employ, e.g., Korean has one word that
represents both green and blue, the development of color terminology and perception
is subject to universalist constraints due to biology \cite{berlin1969basic}.
Recent years, however, have witnessed a resurgence of a milder
strain of the hypothesis, alternatively known as neo-Whorfianism or
the weak Sapir--Whorf hypothesis \cite{boroditsky2003linguistic}. One prominent
controversial research direction investigates this hypothesis with respect to grammaticalized
notions of gender and tense.  Based on experimental evidence obtained from native German and Spanish speakers,
\newcite[\S 4.6]{boroditsky2003linguistic} 
argued that grammatical gender
affects how speakers view objects in their native language.
Similarly, 
differences in the usage of tense in Mandarin and English were found to
affect how the respective speakers 
view time \cite{Boroditsky01doeslanguage}.  In this
work, we focus on the influence of grammatical gender and how we can use 
 NLP tools
to shed light on the validity of this aspect of neo-Whorfianism.

The gist of \newcite[\S 4]{boroditsky2003sex}'s hypothesis is
that speakers of languages that mark grammatical gender perceive
nouns, even inanimate ones, differently depending on the noun's
gender. They sought psycholinguistic evidence for this claim, arguing
that
speakers will more often choose stereotypically masculine adjectives
to describe grammatically masculine inanimate nouns and stereotypically feminine
adjectives to describe grammatically feminine inanimate nouns. To give a
concrete example, the German word for key, \word{Schl{\"u}ssel}, happens
to be masculine, so speakers are more likely to use words such as
\word{heavy}, \word{jagged}, and \word{hard}. In contrast, the Spanish
word for key, \word{llave}, happens to be feminine, so speakers are more
likely to use stereotypically feminine words such as
\word{elegant}, \word{pretty}, and \word{delicate}.  However, \newcite{mickan2014key} failed to replicate this experiment,
leaving the validity of the findings questionable.

In this work, we provide evidence against \newcite[\S 4]{boroditsky2003sex}'s hypothesis using NLP techniques. We contend that if this
instance of neo-Whorfianism is
true, then we should see a reflection of it in corpus co-occurrence
counts. We propose two experiments, based on word embeddings
trained on large corpora, to investigate how speakers of different
languages use certain nouns.  Our results on 9 languages
suggest that the grammatical gender of inanimate nouns does not
influence how they are used in context, taking credence away from
\newcite[\S 4]{boroditsky2003sex}'s claims. However, we
caution that it is important not to overstate the findings of one
study; we see our results as providing further evidence in the
linguistic relativity debate against neo-Whorfianism from a
largely orthogonal source.

\section{Background}\label{sec:background}
\subsection{Grammatical Gender}\label{sec:grammatical-gender}
In this section, we provide a brief overview of grammatical gender
systems since those play an important role in \newcite[\S 4]{boroditsky2003sex}'s, and hence our, experiments.
Languages range from encoding no grammatical gender on
inanimate nouns,
like English or Mandarin Chinese,
to distinguishing tens of gender-like noun classes, as found
in the Bantu languages of Africa \cite{corbett1991gender}.
In this work, we will exclusively consider gendered languages
of Indo-European and
Afro-Asiatic (Semitic) stock; they all distinguish either two or three
genders: the bipartite distinctions masculine-feminine and
the tripartite distinction masculine-feminine-neuter.
An important assumption of our study is the tenet 
that the gender assigned to inanimate objects is arbitrary.

All languages we 
experiment on exhibit \textbf{concord} in
grammatical gender, i.e., articles and adjectives that modify a noun
must agree with that noun in gender as well as in other
features. Consider the German and Spanish translations of the English
sentence: \textit{The beautiful key is on the table}.\arya{Changed your Spanish to make it sound more natural. Word order and adposition change.}
\begin{align}
  &\textit{La llave bonita est{\'a} sobre la mesa.} \\ &\textit{Der
    sch{\"on}e Schl{\"u}ssel liegt auf dem Tisch.}
\end{align}
Here, German employs the masculine article \word{der} as
\word{Schl{\"u}ssel} is masculine; had \word{Schl{\"u}ssel} been
feminine, German speakers would say \word{die Schl{\"u}ssel}. Spanish
exhibits a similar behavior, making use of \word{la}, rather than
\word{el}, as \word{llave} is feminine.  For a thorough treatment of
the subject, see \newcite{corbett2006agreement}.

A key part of our experimental design will be stripping side effects
of the grammatical concord from articles, adjectives, and verbs, thus
removing overt signals of the noun's gender.

\subsection{Morphological Tagging and Lemmatization}\label{sec:lemmatization}
Our study will make use of NLP tools that automate bits of
linguistic analysis: \textbf{morphological tagging} and
\textbf{lemmatization}. Both will be explained here.

In languages that exhibit inflectional
morphology, we may decompose a word into a bundle of morpho-syntactic
features and a lemma, its canonical form.  
Formally, we denote a word in a natural language as $w$, and a sentence of length~$n$ as
$\mathbf{w} = w_1 \cdots w_i \cdots w_n$.  Each word can be factored
into a lemma~$\ell$ and a bundle of morphological features~$m$. For
instance, we may think of the German word \word{Schl{\"u}ssels}
as the lemma $\ell = \word{Schl{\"u}ssel}$ and the
 features $m =
\left[\right.\textsc{pos}$$=$$\textit{n},\enskip \textsc{gen}$$=$$\textit{masc},\enskip \textsc{num}$$=$$\textit{sg}, \enskip\textsc{case}$$=$$\textit{gen}\left.\right]$. Each
morphological feature in $m$ is an attribute--value pair.
Attributes encompass lexical properties such as gender, number, and case,
taking values such as masculine, singular and genitive,
respectively.
Following \newcite{booij1996}, we may divide the morphological
attributes into two categories: \textbf{inherent} and
\textbf{contextual}.  Inherent categories are those that are embedded
in the lemma itself: the lemma \word{Schl{\"u}ssel}
without any additional inflection reveals $\textsc{pos}$$=$$\textit{n}$
and $\textsc{gen}$$=$$\textit{masc}$.  However, the sentential
context dictates that
$\textsc{num}$$=$$\textit{sg}$ and $\textsc{case}$$=$$\textit{gen}$.
We write $\mathbf{m}$ for a sequence of morphological tags and
${\boldsymbol \ell}$ for a sequence of lemmata.
A morphologically tagged Russian example sentence is given in \cref{fig:sentence}.

In general, each word in a sentential context will have exactly one lemma
and one bundle of morpho-syntactic attributes.
We may think of a decomposed sentence of length $n$ as an interleaved
trisequence: $\langle w_1, \ell_1, m_1\rangle \cdots \langle w_n, \ell_n, m_n\rangle$,
where $w_i$ is the $i^\text{th}$ word, $\ell_i$ is its lemma, and $m_i$ 
is the set of its morphological features. 

Several techniques exist 
to map sentences to the lemmata of their words together with their morpho-syntactic attributes.
The task of mapping a sentence to a sequence
of morphological tags, in above notation $\mathbf{w} \mapsto \mathbf{m}$,
is known as morphological tagging. The task
of mapping a sequence of words to a sequence of lemmata, i.e.,
$\mathbf{w} \mapsto {\boldsymbol \ell}$, is known as
lemmatization. Performing the tasks jointly can improve
performance \cite{muller-EtAl:2015:EMNLP}.

\begin{table}
  \centering
  \small
  \setlength{\tabcolsep}{3.0pt} 
  \begin{adjustbox}{width=\columnwidth}
  \begin{tabular}{c  c c c c c c c c c c c c c c c c c c c c c c} \toprule
  \vspace{.1cm} & bg & es & fr & he 
  & it & pl & ro & ru & sk  \\ \midrule
all & 96.6  & 98.5 & 98.4 
& 96.5 & 98.2 & 96.5 & 98.1 & 96.5 & 95.0  \\
OOV & 82.1 & 91.7 & 89.6  & 79.6 
& 91.2 & 86.9 & 89.0 & 88.1 & 86.6 \\
  \bottomrule
  \end{tabular}
  \end{adjustbox}
  \caption{Token-level lemmatization accuracy obtained by \textsc{lemming} on the test splits of the UD treebanks for all languages when evaluated on all tokens or just OOVs.}
  \label{tab:lemming}
\end{table}

\subsection{Word Embeddings}\label{sec:word-embeddings}
In our experiments, we will make strong use of embeddings of words
into $\mathbb{R}^d$. Given a fixed vocabulary 
$V = \{v_1, \ldots, v_{|V|}\}$ of word types,
we will denote the embedding of a type $v$ as
$\mathbf{e}(v) \in \mathbb{R}^d$.\footnote{For notational clarity, we
  use different symbols for types and tokens. In this work, $v_i$ will
  always denote a word type, that is, the $i^\text{th}$ lexical item
  in a fixed vocabulary $V$, whereas $w_i$ will denote the
  $i^\text{th}$ token in a sentence $\mathbf{w}$.}
  
We employ the
  \textsc{word2vec} \cite{mikolov2013efficient,goldberg2014word2vec}
  toolkit, in particular the skip-gram model, 
  for the creation of our word embeddings. Skip-gram
  may be considered a form of matrix factorization; specifically,
  it factorizes a matrix of probabilities $X \in \mathbb{R}^{|V| \times |V|}$, where $X_{ij}$
  denotes the probability that $v_i$ co-occurs with $v_j$ within a certain context
  window. For instance, a symmetric context window of size 5
  is a common choice. This 
  asks how often $v_j$ occurs within five positions to the left or right
  of $v_i$. Under this interpretation, \textsc{word2vec} is an instance
  of exponential-family PCA
  \cite{DBLP:conf/nips/CollinsDS01,cotterell-et-al-2017-eacl}. The
  output of \textsc{word2vec} is a mapping of word types to a
  vector space: $\mathbf{e} : V \rightarrow \mathbb{R}^d$.  
  We highlight that the model only
  considers words categorically, i.e., it is unable to look at the
  surface form of the word. This is relevant as there are
  sub-word indicators of gender, e.g., feminine nouns in
  Spanish often end in \word{-a}.

In practice, embeddings are taken as a proxy for lexical semantic
meaning---words that have similar meanings should be closer together
in the space.  This idea has a long history in NLP;
\newcite{firth1957synopsis} famously quipped, ``You shall know a word
by the company it keeps.'' As we will discuss in \cref{sec:original},
are we interested in how well the embeddings for \emph{inanimate
  nouns} encode the respective genders.

\section{Neo-Whorfianism and Word Embeddings: What is the Link?}\label{sec:original}

Our primary contribution is the development of a computational analogue of the previously conducted
psycholinguistic study by \newcite[\S 4]{boroditsky2003sex}, mentioned in \cref{sec:introduction}. While,
naturally, the signals in a big-data analysis are different
than those extracted from subjects in a laboratory, we find it useful
to first describe the original work. 

\subsection{The Psycholinguistic Experiment}\label{sec:psycho}
To test the hypothesis
that the
grammatical gender assigned to inanimate objects, even though
it is not (and, in fact, \emph{cannot} be) a
direct reflection of natural gender (since the latter is not defined for inanimate nouns),
has an influence on
the manner in which speakers perceive those objects, \newcite[\S 4]{boroditsky2003sex} use the
following experimental procedure. They created a list of 24 words in
German and Spanish that were selected to be translations of each other. Importantly,
the same number of masculine and feminine nouns were present in each
language; however, all nouns had different genders in German and
Spanish. We show a list of stimuli for the experiment in \cref{tab:stimuli},
taken from \newcite[Appendix A]{Boroditsky00sex}.
Then, native speakers of each language were brought into the
laboratory and asked to describe the nouns with the first three
adjectives that came to mind. (Note the that the experiment took place
in English, even though each subject's native language was either German
or Spanish.) 
As a second step, a group of native English
speakers were asked to judge the elicited adjectives as either $-1$
(feminine) or $+1$ (masculine), which yielded a gender rating. Using
this rating, \newcite[\S 4]{boroditsky2003sex} found a correlation
between the genderedness of the German and Spanish speakers' choice of
adjectives in the first experiment and the English speakers' rating of
how gendered each adjective was. This was taken to be a very
conservative test for neo-Whorfian effects regarding gender,
as the
experiment took place in English, which lacks grammatical gender.

As a qualitative example, they report that the German-speaking
participants described ``key,'' which is masculine in German, as
\word{hard}, \word{heavy}, and \word{jagged}, whereas the Spanish-speaking
participants described ``key,'' which is feminine in Spanish, as \word{beautiful},
\word{elegant}, and \word{fragile}. They argue that these findings show
that the manner in which German and Spanish speakers think about
inanimate objects is influenced by the grammatical
gender that the language assigns to their nouns.

\begin{table}
  \centering
  {\small
  \begin{tabular}{l ll ll} \toprule
    \textbf{English} & \multicolumn{2}{c}{\textbf{Spanish}} & \multicolumn{2}{c}{\textbf{German}} \\ \cmidrule(l){1-1} \cmidrule(l){2-3} \cmidrule(l){4-5}
    \word{apple} & \word{manzana} & f. & \word{Apfel} & m. \\
    \word{arrow} & \word{flecha} & f.  & \word{Pfeil} & m. \\
    \word{boot}  & \word{bota} & f.    & \word{Stiefel} & m. \\
    \word{broom} & \word{escoba} & f.  & \word{Besen} & m. \\
    \word{moon} & \word{luna} & f. & \word{Mond} & m. \\
    \word{spoon} & \word{cuchara} & f. & \word{L{\"o}ffel} & m. \\
    \word{star} & \word{estrella} & f. & \word{Stern} & m. \\
    \word{toaster} & \word{tostador} & m. & \word{R{\"o}ster} & m. \\
    \word{pumpkin} & \word{calabaza} & f. & \word{K{\"u}rbis} & m. \\ \midrule
    \word{bench} & \word{banco} & m. & \word{Bank} & f. \\
    \word{brush} & \word{cepillo} & m. & \word{B{\"u}rste} & f. \\
    \word{clock} & \word{reloj} & m. & \word{Uhr} & f. \\
    \word{disk} & \word{disco} & m. & \word{Scheibe} & f. \\
    \word{drum} & \word{tambor} & m. & \word{Trommel} & f. \\
    \word{fork} & \word{tenedor} & m. & \word{Gabel} & f. \\
    \word{sun} & \word{sol} & m. & \word{Sonne} & f.  \\
    \word{toilet} & \word{inodoro} & m. & \word{Toilette} & f. \\
    \word{violin} & \word{viol{\'i}n} & m. & \word{Geige} & f. \\
    \bottomrule  \end{tabular}
  }
  \caption{A list of 18 English words for common inanimate objects with their translations
    into both Spanish and German. The list, inspired by \newcite[Appendix A]{Boroditsky00sex}, demonstrates
  that grammatical gender is relatively arbitrary in how it differs between languages. }
  \label{tab:stimuli}
\end{table}

The findings of \newcite[\S 4]{boroditsky2003sex} have not gone
uncontested.  \newcite{mickan2014key} report two unsuccessful attemps
to replicate the experiments described in \cref{sec:original}. They also
note that, while the study is widely cited, the experiments, along with
their experimental stimuli, were never
published in their own right, but rather merely described in a summary
book chapter.  Nevertheless, the idea that grammatical gender may
influence thought has taken off with much ink spilled on the subject
in the popular press.  Indeed, in partial response to the popularity of
the idea, \newcite{mcwhorter2014language} authored an entire volume, \textit{The
  Language Hoax}, with the explicit purpose of removing much of the
hype surrounding neo-Whorfian claims.

\subsection{Neo-Whorfianism, Gender, and Word Embeddings}\label{sec:neo-whorphianism-gender-embeddings}
\Ryan{Revisit introduction after rewriting 3.2. And the conclusion as well.}
How can we use NLP to test the claims investigated in \cref{sec:psycho}?
We contend that the thrust of \newcite[\S 4]{boroditsky2003sex}'s argument may be reduced to
one of basic lemma co-occurrence counts in a large corpus.
  If German
speakers are more likely to describe a \word{Schl{\"u}ssel}
(\word{key}) with a stereotypically masculine adjective, i.e., \word{jagged},
rather than with a stereotypically feminine one, i.e., \word{delicate}, we
should reasonably expect this predisposition to manifest itself in
corpus counts: grammatically masculine nouns should
have stereotypically masculine adjectives modifying them with higher frequency, while
feminine nouns should be more likely to be modified by stereotypically feminine ones.
\katha{Note here that this is caused by it being the first one that
  comes to people's minds... like in Boroditsky's experiments.}
\lawrence{Just a thought: Wikipedia may be a domain in which the
  notion of superfluous adjectives to describe nouns (which
  Boroditsky's study exploits) doesn't really occur, since it is
  factual enciclopedia style, and neither flowery imagery nor personal
  opinions are well represented.}  \ryan{Not sure domain matters, if
  it is a subconscious effect}

Linking this idea to NLP, recall from \cref{sec:word-embeddings} that
co-occurrence counts are the primary signal for training word
embeddings, one of the most popular lexical semantic representations
of meaning at the type level.  Expanding on
\newcite{firth1957synopsis}'s original statement, we further ask if
you shall also know the word's \textit{grammatical gender} from its
kept company? Operationalizing this, we will attempt to predict the
gender of a noun type from its word embedding under several
experimental conditions. This is a reformulation of \newcite[\S
  4]{boroditsky2003sex}'s experimental paradigm. The original
experiment asked the participants to generate adjectives given a noun
stimulus, whereas we look at the spontaneously written contexts of nouns in large corpora.

In more detail,
\newcite[\S 4]{boroditsky2003sex}'s participants
are given nouns as stimuli, whose gender they assume the participants have access to in their internal representation
of the lexicon. (Recall that gender is an inherent
morphological property, as discussed in
\cref{sec:grammatical-gender}.) Then, those participants are
asked to generate adjectives that would be good
descriptors for those nouns, i.e., they generate contexts for those
nouns. The contexts are
then scored in a second experiment, where
English speakers determined how gender-stereotyped
the adjectives are. 
In our experiments, we induce a representation of the noun's
context from copious amounts of raw text. (The
context will include adjectives, like the ones
\newcite[\S 4]{boroditsky2003sex}'s participants generate.)
Then, we employ a classifier to
reconstruct the gender of the inanimate noun from its embedded
(lemmatized) context. 
Despite this difference, the idea tested is identical:
can one predict a noun's grammatical gender from the words humans associate with it?

Methodologically, we will introduce two experiments
that investigate the relation between a noun's grammatical gender and the noun's context. Experiment 1
(\cref{sec:experiment1}) attempts to predict the gender
from a word embedding with a black-box classifier, as we hint at above. 
Experiment 2 (\cref{sec:experiment2}) complements
experiment 1 in that it provides a technique for analyzing the ``genderedness'' of inanimate
nouns. Both experiments make used of the same multilingual
data, whose preparation is described in \cref{sec:preparation}.

\section{Experimental Setup}\label{sec:preparation}
The goal of both our experiment and that
of \newcite[\S 4]{boroditsky2003sex} is to determine
whether the words that occur in the context of a noun
are influenced by its grammatical gender.
In our experiment, we opt to represent
a context by its word embedding and try to predict the gender of a (inanimate) noun given its word embedding.

Extracting the gender of a word from its vector representation is
often trivial for many languages due to certain grammatical artifacts:
e.g., in Spanish, nouns are usually accompanied by a gender-specific
article (\word{el} or \word{la}). Thus, in order to obtain meaningful
experimental results, we need to control for such obvious indicators,
i.e., lemmatize our corpora and train lemma embeddings instead of
embeddings for all inflected forms.  Indeed, word embeddings famously
capture gender, as evinced by
\newcite{mikolov-yih-zweig:2013:NAACL-HLT}'s (approximate) equation
\begin{equation*}
\mathbf{e}(\text{\word{king}}) - \mathbf{e}(\text{\word{man}}) + \mathbf{e}(\text{\word{woman}}) \approx \mathbf{e}(\text{\word{queen}}). 
\end{equation*}
Recall from \cref{sec:word-embeddings} that our embeddings
do not have access to subword information, so clues
for a noun's gender \emph{must} come from context.

\subsection{Word Embedding Comparison}
We induce word embeddings under four experimental conditions: (i) \textbf{forms}: the embeddings are
trained on forms (the original corpus), (ii) \textbf{lemmata}: the
embeddings are trained on lemmata (the whole corpus is lemmatized),
(iii) \textbf{nouns}: the embeddings are trained on lemmatized nouns
where the rest of the corpus is left unlemmatized, (iv)
\textbf{$\neg$nouns}: the embeddings are trained on unlemmatized nouns
and the rest are lemmatized.

\paragraph{Hypotheses.}
We compare the ability of the classifier to predict the gender of the
noun in the word conditions outlined above and, additionally, compare
the results to a majority-class baseline.  We hypothesize the
classifier to easily be able to predict the gender from conditions (i) \textbf{forms}
and (iii) \textbf{nouns} since the primary cue for gender is the concord exhibited by
the context words.  Indeed, we see no \textit{a-priori} reason why (i)
should perform significantly differently than (iii). The conditions (ii) \textbf{lemmata} and (iv)
\textbf{$\neg$nouns} are more interesting: If the inherent gender in the inanimate nouns
influences the choice of context lemmata, as
\newcite[\S 4]{boroditsky2003sex} believe, then we hypothesize the classifier
to be able to predict the gender from the embeddings in (ii) \textbf{lemmata} and (iv)
\textbf{$\neg$nouns}
better than a majority-class baseline. However, if
speakers are uninfluenced by grammatical gender, then we should
fail to predict grammatical gender from context. We note (i) and
(iii) are skylines since (ii) and (iv) contain less
gender-related information.
 
\subsection{Data Preparation}\label{sec:preparation}
\Ryan{Hindi numbers are a place holder}
The first step in the data creation pipeline is lemmatization. This is necessary to remove grammatical concord, for
instance, because the Spanish phrase \word{la llave bonita} shows
concord between the feminine noun \word{llave}, the definite article
\word{la} and the adjective \word{bonita}. The lemmatized version of
the sentence is \word{el llave bonito}, where both the article and the
adjective have been reverted to their canonical form, which,
traditionally, is the masculine singular form in Spanish. Hence,
gender information is no longer trivially detectable using neighboring
words.

\paragraph{Lemmatizer.}
We choose the \textsc{lemming} \cite{muller-EtAl:2015:EMNLP}
package\footnote{\url{http://cistern.cis.lmu.de/lemming/}} for
lemmatization. \textsc{lemming} is a conditional random field
\cite{lafferty2001conditional} with artisanal feature templates. While
occasionally surpassed in performance, \textsc{lemming} is competitive
with neural state-of-the-art models, while remaining robust in
low-resource settings and being fast to train
\cite{heigold-neumann-vangenabith:2017:EACLlong}.  For all languages
in our experiments, we keep \textsc{lemming}'s defaults as our
hyperparameters and train it on the training and development splits of
the corresponding Universal
Dependencies (UD) treebanks \cite{11234/1-1983}.
Since errors during lemmatization could significantly alter our
experiments, we make use of the dev splits to ensure that a high
lemmatization performance is reached. \cref{tab:lemming} shows the
resulting token-based accuracies of our final models, both on the
entire UD test sets and on out-of-vocabulary words only. The latter
serves as a lower-bound; it corresponds to the (extreme) case where
none of the words of the corresponding Wikipedia corpus appear in the
union of the UD training and development sets.

\paragraph{Experimental Languages.}
We choose 9 experimental languages randomly from UD: Bulgarian (bg),
French (fr), Hebrew (he), Italian (it), Polish (pl),
Romanian (ro), Russian (ru), Slovak (sk), Spanish (es). The only
imposed condition is that that they have $\geq 80\%$ performance on
lemmatizing out-of-vocabulary words on the UD development set. This is
to ensure that our lemmatizer is reasonably leak-free for rare words.
We drop neuter words in languages that exhibit a neuter,
such as Bulgarian and Russian.

\paragraph{Word Vectors.}
We employ the skip-gram model from the \textsc{word2vec} package to
induce 100-dimensional embeddings. We use negative sampling with 10
samples. For all languages, the vectors are trained on corpora
lemmatized in the way we just described; namely, we make use of
multilingual Wikipedia editions from March 2018.  All words with a
frequency below $5$ are ignored, and we compare symmetric context window sizes
of $2$, $5$ and $10$, finding $2$ works the best.

\subsection{Experiment 1: Gender Classification}\label{sec:experiment1}
Our gender prediction problem constitutes a binary
classification task, where the classes
are masculine and feminine and the input is the 
word embedding of a noun. We employ a multi-layer perceptron (MLP) for
this classification, defining the probability of the gender $g$
given a noun type $v$, as
\Ryan{Equation 3 is still ugly to save space. Hanna will expect a change in the camera ready.}
\begin{align}\label{eq:mlp}
&p(g \mid v) =\notag\\
&\quad\text{softmax}(W_2 \tanh \left(W_1 \mathbf{e}(v) +
b_1\right) + b_2) 
\end{align}
where we feed in the noun's embedding $\mathbf{e}(v)$ into the network, $W_1 \in
\mathbb{R}^{d' \times d}$ and $W_2 \in \mathbb{R}^{2 \times d'}$ are
weight matrices, $b_1 \in \mathbb{R}^{d'}$ and $b_2 \in
\mathbb{R}^{2}$ are bias vectors. \Cref{eq:mlp} represents a
network of depth 2, but we consider depth-$k$ networks
where $k$ is a hyperparameter. We additionally consider the non-linearities $\textit{ReLU}$
and $\textit{sigmoid}$.

\begin{figure}
\centering
\includegraphics[width=\columnwidth]{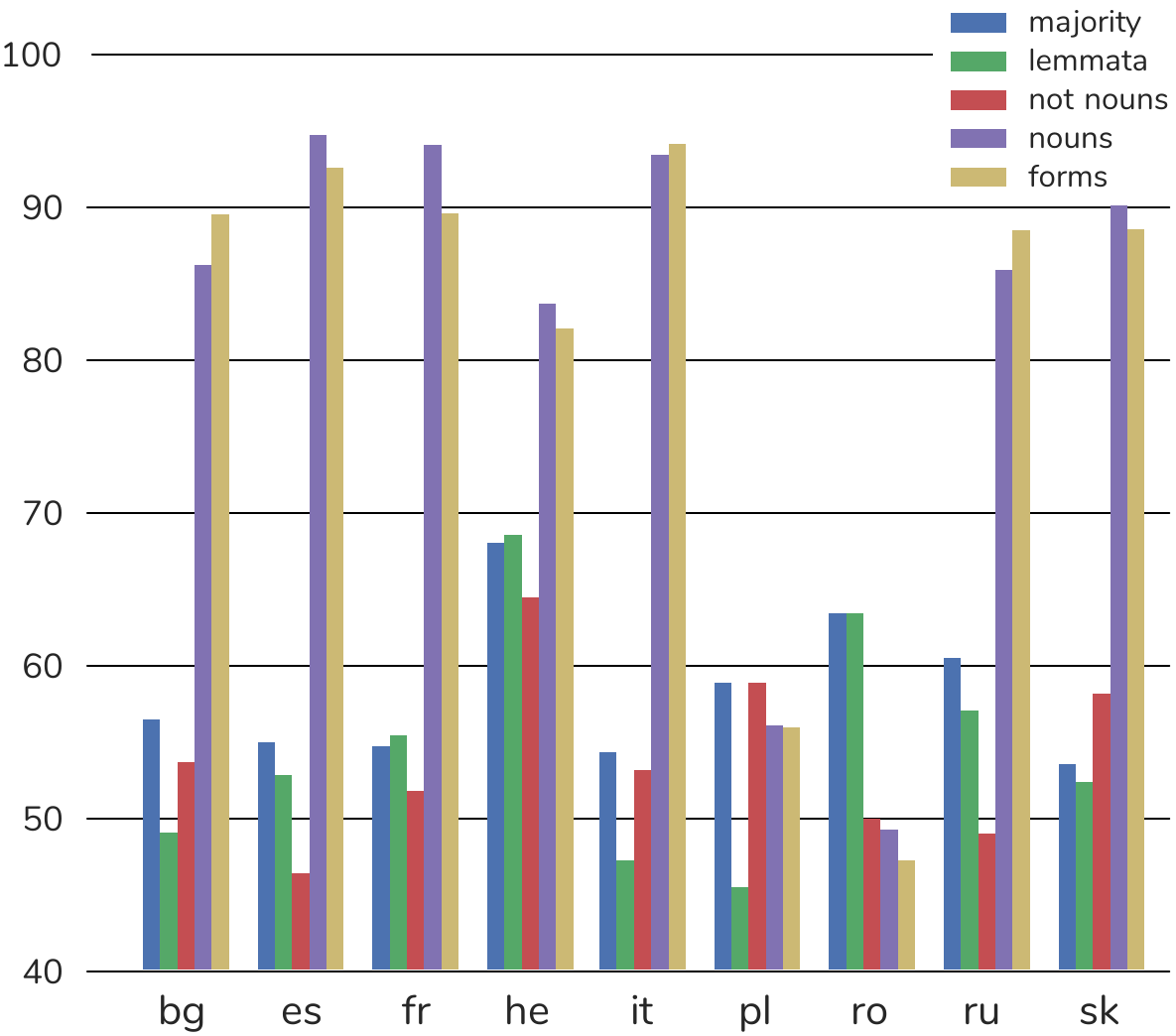}
\caption{Accuracies for our classifer on all languages. The
  majority-class baseline (blue bar) is generally above the lemmata-based prediction (green bar), but
  much below the form-based prediction (yellow bar). This leads us to conclude that we \emph{cannot} reliably
predict grammatical gender for inanimate nouns from context alone.}
\label{fig:gender_results}
\end{figure}
\paragraph{Training Set.}
To learn the parameters of \cref{eq:mlp}, we construct
the following training set. Given the lemmatized and morphologically analyzed Wikipedias
discussed in \cref{sec:preparation}, we construct a
lexicon as follows. For every lemma type that occurs
more than 50 \ryan{Double check!} times,
we find the gender, extracted from the morphological 
tag of each token, that most frequently occurs among its tokens
in the corpus. This yields a lexicon of lemma-gender pairs. 
Note that this training set will include animate words as we are
unable to exclude them easily. However, we will only evaluate on inanimate nouns; see below.

\paragraph{Evaluation Set.}
For the evaluation, we focus exclusively on the same common, inanimate
words in all languages. We use the NorthEuraLex dataset \cite{northeuralex},\footnote{\url{http://www.northeuralex.org/}} which is a multi-way
concept-aligned dictionary. In order to avoid biological gender interfering with
grammatical gender and, thus, influencing our experiments, we manually
annotate all concepts as animate or inanimate and exclude all animate
nouns. For instance, we keep \textit{eye}, \textit{lake}, and
\textit{circle}, while discarding words like \textit{wife},
\textit{dog}, or \textit{son}.  The list containing all concepts in
our evaluation set can be found in \cref{sec:northeuralex}. We take care to 
remove these words from the training set. Note that not all words are observed
in each language, due to the varying sizes of the Wikipedia corpora.
We split the resulting lexicons in half, creating a dev
and test set for each language.

\paragraph{Training and Hyperparameters.}
The model is implemented in PyTorch \cite{paszke2017automatic}. 
We train our models on the 
training sets using Adam \cite{KinBa17} with a base learning
rate of $0.1$ for all models. All models are trained for $50$ epochs.
Hyperparameters include network depth ($k \in [1, 5]$),
size of the hidden layer (taken from $\{100, 200, 300\}$) and
type of nonlinearity (taken from $\{\tanh, \textit{sigmoid}, \textit{ReLU} \}$). We randomly partition the evaluation
set in half 10 times and sweep the hyperparameters on each dev partition,
performing early stopping.
Final results average the performance on test across these splits.
\begin{table}
  \small
  \centering
  \begin{adjustbox}{width=\columnwidth}
  \setlength{\tabcolsep}{5.pt}
  \begin{tabular}{ll ll} \toprule
    \multicolumn{2}{c}{\textbf{forms}} & \multicolumn{2}{c}{\textbf{lemmata}}\\ \cmidrule(l){1-2} \cmidrule(l){3-4}
    \textbf{fem} & \textbf{masc} & \textbf{fem} & \textbf{masc} \\ \cmidrule(l){1-2} \cmidrule(l){3-4}
  \word{estrella} (f.) & \word{sonido} (m.) & \word{labio} (m.) & \word{hoguera} (f.) \\
  \word{ola}  (f.) & \word{labio} (m.) & \word{mejilla} (f.) & \word{aldea} (f.) \\
  \word{bolsa}  (f.) & \word{tejido} (m.) & \word{entendimiento} (m.) & \word{esqu{\'i}}  (m.)  \\
  \word{isla}  (f.)  & \word{lazo} (m.) & \word{{\'a}ngulo} (m.) & \word{altitud} (f.) \\
  \word{carta}  (f.)  & \word{nudo} (m.) & \word{vena} (f.)  & \word{silla} (f.)  \\
  \bottomrule
  \end{tabular}
  \end{adjustbox}
  \caption{The most gendered nouns in our Spanish test set according to our
    induced gender dimension under the \textbf{forms} and the \textbf{lemmata} conditions. The
    gender dimension is the column header and the true gender is marked in parentheses. \label{tab:most_gender}}
\end{table}

\paragraph{Results and Discussion.}
\Ryan{Tie it back even more to the hypothesis section} The
gender-prediction accuracies from the four embeddings conditions are
shown in \cref{fig:gender_results}.
As discussed, classification accuracy
is highest for conditions (i) \textbf{forms} and (iii) \textbf{nouns}, i.e., embeddings trained on
the corpora where the context words are unaltered. In these
conditions, where we see unlemmatized forms as context, our classifier
handily surpasses the majority-class baseline with differences up to
30 points.  All differences are significant ($p < 0.05$). On the other
hand, inspecting conditions (ii) \textbf{lemmata} and (iv) $\neg$\textbf{nouns}, we see that performance of
each is rarely better than the majority-class baseline and in \emph{no
  case} is statistically better ($p < 0.05$). Thus, despite a
relatively extensive hyperparameter search, we are \emph{unable} to
reliably predict grammatical gender from the context words along.
This negative result provides evidence \emph{against} \newcite[\S
  4]{boroditsky2003sex}'s hypothesis that the inherent gender of the
word will have an effect on the context words that a speaker uses for
inanimate nouns.

\Arya{Isn't 0.05 a bit generous?}

\subsection{Experiment 2: A Gender Dimension}\label{sec:experiment2}
In addition to Experiment 1, we would also like to analyze
the degree to which words are more masculine or feminine using their word embeddings, which, in turn,
tells us how masculine or feminine their contexts
are.
A high-level overview of how we can achieve this is as follows.
We may isolate genderedness by fixing one of the dimensions
of the word embeddings to be the gender dimension.
Then, we seek a method that will shift the information regarding gender into that dimension.
In effect, we need a method which maps every word to a scalar quantity which corresponds to how gendered the word is, allowing us to compare individual words and to discover those whose gender is more saliently encoded in the embeddings. We adopt the \textit{ultra-dense} strategy developed by \newcite{rothe-ebert-schutze:2016:N16-1}, which we will describe below. 

\paragraph{Learning Ultra-Dense Embeddings.}
Given an embedding $\mathbf{e}(v) \in \mathbb{R}^d$ of a word $v$, we are interested in
learning a real orthogonal matrix $Q \in \mathbb{R}^{d \times d}$ in
order to create a new embedding $\mathbf{e}'(v) = Q\, \mathbf{e}(v)$.
Defining $Q$ to be real orthogonal ensures that 
no information is lost or gained as a result of the transformation---the dot product, and, thus, the cosine similarity
between vectors will be preserved.
In order to learn a transformation that moves gender information to certain components of the embeddings, 
let ${\cal S}$
be the set of of all pairs of distinct nouns that have the \emph{same} gender and let ${\cal D}$ be the
set of all pairs of distinct nouns that
have a different gender. Let $P \in \mathbb{R}^{d \times d}$ be a matrix with all entries being zero
except for $P_{11}$, which is 1. Now, we minimize the following objective
\begin{align}
  \mathcal{O}\Big(Q; \mathcal{S}, &\mathcal{D}\Big) = \!\!\!\sum_{(v, v') \in {\cal S}} \!\!\! \mid\mid P\, Q\left(\mathbf{e}(v) - \mathbf{e}(v')\right)\mid\mid^2_2 \nonumber  \\
     -\!\!\!&\sum_{(v, v') \in {\cal D}} \!\!\! \mid\mid P\, Q\left(\mathbf{e}(v) - \mathbf{e}(v')\right)\mid\mid^2_2  \label{eq:obj_fct}
\end{align}
with respect to the matrix $Q$ subject to the constraint that $Q^{\top} Q = I$; that is, $Q$ is real orthogonal.

\paragraph{Stochastic Projected Gradient.}

The above objective can be optimized using a stochastic
projected-gradient-style algorithm \cite{bertsekas1999nonlinear}.  This algorithm
alternates between two steps until convergence: (i) A stochastic gradient
step: During this step, one element is randomly sampled from each of ${\cal S}$
and ${\cal D}$. Then, the gradient of
\cref{eq:obj_fct} is computed with respect to $Q$, and $Q$ is 
updated by taking a step in the direction of the gradient. (ii) A projection step: After obtaining a new matrix $Q' = Q + \eta \cdot \nabla_Q {\cal O}$
during the gradient update where $\eta$ is the learning rate, we
no longer have the guarantee that $Q'$ is orthogonal. Thus, we must perform a projection step to orthogonalize
$Q'$. This can be achieved through singular value decomposition (SVD).  We compute the SVD: $Q' = U \Sigma V^{\top}$, where $U$,
$\Sigma$ and $V^{\top}$ are guaranteed to be real as $Q'$ is real. Then, we may define
$Q = UV^{\top}$, which is the closest to $Q'$ under the Frobenius norm
\cite{horn1990matrix}. Pseudocode for this algorithm is given 
in \cref{alg:spga}.

\begin{algorithm}[t]
  \begin{algorithmic}[1]
    \State{\textbf{input} $\mathcal{S}, \mathcal{D}$ \Comment{\textit{same- and different-gender pairs}}}
    \State{$Q \gets I$}
    \For{$t=1$ \textbf{to} $T$}
    \State $(v_{\cal S}, v_{\cal S}') \sim \text{uniform}({\cal S})$
    \State $(v_{\cal D}, v_{\cal D}') \sim \text{uniform}({\cal D})$
    \State $\tilde{{\cal S}} \gets \set{(v_{\cal S}, v_{\cal S}')}; \tilde{{\cal D}} \gets \set{(v_{\cal D}, v_{\cal D}')}$
    \State{$Q' \gets \eta_t \cdot  \nabla_Q \, \mathcal{O}(Q; \tilde{\mathcal{S}}, \tilde{\mathcal{D}})$} 
    \State $U\Sigma V^{\top} \gets \texttt{SVD}\left(Q'\right)$
    \State $Q  \gets UV^{\top}$
    \EndFor
  \end{algorithmic}
  \caption{{\small Stochastic Projected Gradient Algorithm}}
  \label{alg:spga}
\end{algorithm}

\paragraph{Training Details.}
Our learning rate schedule $\eta_t$ (see \cref{alg:spga}) is chosen by
the Adam optimizer \cite{KinBa17}. We run $T=1000$ iterations. Upon
termination, we extract a scalar-valued gender quantity as
follows: $\left[P\, Q\, \mathbf{e}(w)\right]_{11}$, i.e., the first
component of the new embedding. We train on half the NorthEuraLex data and test
on the other half, using the splits described in \cref{sec:experiment1}.

\paragraph{Analyzing the Data.}
The gender dimension admits both a quantitative and a qualitative
analysis. We start with a quantitative analysis; we consider
Spearman's $\rho$ between the gender dimension and the grammatical
gender of the nouns, marking masculine as 0 and feminine as 1. The
results are shown in \cref{table:corr}.  They mirror those found in
experiment 1 (\cref{sec:experiment1}): we are unable to find
correlation significantly different than 0 for any of the cases where
the context words have been lemmatized (conditions (i) \textbf{lemmata} and (iii) $\neg$\textbf{nouns}). On
the contrary, when the context words are left unlemmatized (conditions (i) \textbf{forms} and (iii) \textbf{nouns}), we are generally able to find a significant
correlation. Qualitatively, the gender dimension tells us which words
are more gendered than others. Here, we perform a case study of our
Spanish test set.  The five words with the most masculine and most feminine gender dimension are displayed in
\cref{tab:most_gender} in the (i) \textbf{forms} and (ii) \textbf{lemmata}
conditions. The qualitative analysis shows the same trend.

\Arya{Are you sure table 3 isn't backwards? All your masculine things are feminine and vice versa.}

\setlength{\belowcaptionskip}{-10pt}
\begin{table}
  \small
  \centering
  \begin{tabular}{c  c c c c} \toprule

  {\bf Lang}&\multicolumn{4}{c}{\bf Spearman's $\rho$} \\ \cmidrule(l){1-1} \cmidrule(l){2-5} 
  & \scriptsize{lemmata} & \scriptsize{$\neg$nouns} & \scriptsize{nouns} & \scriptsize{forms}  \\ \midrule
  bg & \textcolor{brickred}{07.80} & \textcolor{brickred}{02.38} & \textcolor{darkmidnightblue}{78.29} & \textcolor{darkmidnightblue}{84.65} \\
es & \textcolor{brickred}{06.80} & \textcolor{brickred}{22.65} & \textcolor{darkmidnightblue}{84.79} & \textcolor{darkmidnightblue}{85.47} \\
fr & \textcolor{brickred}{06.88} & \textcolor{brickred}{19.24} & \textcolor{darkmidnightblue}{84.23} & \textcolor{darkmidnightblue}{85.08} \\
he & \textcolor{brickred}{09.94} & \textcolor{brickred}{18.57} & \textcolor{darkmidnightblue}{56.53} & \textcolor{darkmidnightblue}{63.09} \\
it & \textcolor{brickred}{02.93} & \textcolor{brickred}{09.21} & \textcolor{darkmidnightblue}{80.70} & \textcolor{darkmidnightblue}{82.68} \\
pl & \textcolor{brickred}{14.92} & \textcolor{brickred}{06.46} & \textcolor{brickred}{03.43} & \textcolor{darkmidnightblue}{18.42} \\
ro & \textcolor{brickred}{08.52} & \textcolor{brickred}{14.21} & \textcolor{darkmidnightblue}{53.14} & \textcolor{darkmidnightblue}{82.37} \\
ru & \textcolor{brickred}{18.79} & \textcolor{brickred}{42.02} & \textcolor{darkmidnightblue}{70.09} & \textcolor{darkmidnightblue}{79.22} \\
sk & \textcolor{brickred}{15.14} & \textcolor{brickred}{33.21} & \textcolor{darkmidnightblue}{81.02} & \textcolor{darkmidnightblue}{82.74} \\
\bottomrule
  \end{tabular}
  \caption{\label{table:corr} Spearman's $\rho$ between the generated dimension on test data and the true gender annotation. The correlation is statistically different than zero with $p < 0.05$ in those entries in blue; the ones in red were not found to be significant.}
\end{table}

\section{Other Related Work}
In the realm of NLP, the closest work to ours deals with bias in word
embeddings. Many have observed that word embeddings encode the biases
present in the data they were trained on on. For instance,
\newcite{DBLP:conf/nips/BolukbasiCZSK16} and
\newcite{DBLP:conf/emnlp/ZhaoWYOC17} note that the \word{engineer}
embedding has a higher cosine similarity with \word{man} than with
\word{woman}, reflecting a structural imbalance in the gender of the
profession; they propose to debias the embeddings such that
gender no longer plays a role.

\section{Conclusion}
Using word embeddings in 9 different languages trained on
lemmatized corpora, we investigated whether
adjective choice is influenced by the grammatical gender of inanimate
nouns. This question has larger implication in the debate on the
relation between language and thought. We developed a computational
analogue of \newcite[\S 4]{boroditsky2003sex}'s experimental paradigm
and showed that context in which a noun occurs, stripped of
its overt gender markings, is no longer predictive of the inherent
gender of the original, inanimate noun. These negative results
contradict \newcite[\S 4]{boroditsky2003sex}'s claims.

Any scientific study should be viewed with a healthy dose of
skepticism, especially one, such as ours, that considers a controversial
question. We believe our big-data study should be
taken as complementary evidence in the context of the larger
debate that inanimate nouns'
gender does not influence the way speakers describe them in a corpus.

\section*{Acknowledgments}
I would like to thank Hanna Wallach, Lawrence Wolf-Sonkin, and Ryan Cotterell for discussions and contributions, and they approve this acknowledgment. Except for the addition of author information, these acknowledgments, and minor changes this paper is unchanged from a manuscript written in 2018.

\bibliography{neowhorf}
\bibliographystyle{acl_natbib_nourl}

\cleardoublepage
\appendix

\section{(Inanimate) NorthEuraLex Concepts}\label{sec:northeuralex}

\tablefirsthead{
\toprule
Concept ID & English \\
\midrule
}
\tablehead{
{{\bfseries  Continued from previous column}} \\
\toprule
Concept ID & English \\
\midrule
}
\begin{supertabular}[]{lr}
Abend::N & evening \\
Abhang::N & slope \\
Abstand::N & gap \\
Ader::N & vein \\
Alter::N & age \\
Angelegenheit::N & matter \\
Anh{\"o}he::N & elevation \\
Anzahl::N & count \\
Apfel::N & apple \\
Arbeit::N & work \\
Arm::N & arm \\
Art::N & sort \\
Arznei::N & medicine \\
Asche::N & ashes \\
Ast::N & limb \\
Atem::N & breath \\
Auge::N & eye \\
Bach::N & brook \\
Band::N & ribbon \\
Bart::N & beard \\
Bau::N & lair \\
Bauch::N & belly \\
Baum::N & tree \\
Beere::N & berry \\
Bein::N & leg \\
Berg::N & mountain \\
Besen::N & broom \\
Bett::N & bed \\
Beutel::N & pouch \\
Bild::N & picture \\
Birke::N & birch \\
Blatt::N & leaf \\
Blume::N & flower \\
Blut::N & blood \\
Boden::N & ground, soil \\
Bogen{[}Waffe{]}::N & bow \\
Boot::N & boat \\
Brei::N & mush \\
Brett::N & board \\
Brief::N & letter \\
Brot::N & bread \\
Brunnen::N & well \\
Brust::N & breast, chest \\
Br{\"u}cke::N & bridge \\
Buch::N & book \\
Buchstabe::N & character \\
Bucht::N & cove \\
Busen::N & bosom \\
Butter::N & butter \\
B{\"u}ndel::N & bundle \\
Dach::N & roof \\
Decke::N & blanket \\
Deckel::N & cover \\
Donner::N & thunder \\
Dorf::N & village \\
Dreck::N & filth \\
Ecke::N & corner \\
Ei::N & egg \\
Eimer::N & bucket \\
Eis::N & ice \\
Eisen::N & iron \\
Ellenbogen::N & elbow \\
Ende::N & end \\
Entfernung::N & distance \\
Erde::N & earth \\
Erz{\"a}hlung::N & story \\
Essen::N & meal \\
Faden::N & thread \\
Falle::N & trap \\
Farbe::N & paint \\
Feder::N & feather \\
Fehler::N & mistake \\
Fell::N & fur \\
Fenster::N & window \\
Ferse::N & heel \\
Festland::N & land \\
Fett::N & fat \\
Feuer::N & fire \\
Fieber::N & fever \\
Figur::N & figure \\
Finger::N & finger \\
Fingernagel::N & fingernail \\
Fleisch::N & meat \\
Fluss::N & river \\
Fl{\"u}gel::N & wing \\
Frost::N & frost \\
Funke::N & spark \\
Fu{\ss}::N & foot \\
Fu{\ss}boden::N & floor \\
Gabel::N & fork \\
Gang::N & walk \\
Gast::N & guest \\
Gedanke::N & thought \\
Ged{\"a}chtnis::N & memory \\
Gegend::N & area \\
Gegenstand::N & item \\
Gehirn::N & brain \\
Geist::N & spirit \\
Geld::N & money \\
Gel{\"a}chter::N & laughter \\
Genick::N & nape \\
Geruch::N & odour \\
Geschenk::N & gift \\
Geschirr::N & dishes \\
Geschmack::N & flavour \\
Gesch{\"a}ft::N & business \\
Gesicht::N & face \\
Gespr{\"a}ch::N & talk \\
Gesundheit::N & health \\
Getreide::N & corn \\
Gewalt::N & violence \\
Gewehr::N & gun \\
Gewicht::N & weight \\
Gipfel::N & summit \\
Glas::N & glass \\
Gl{\"u}ck::N & happiness \\
Gold::N & gold \\
Grab::N & grave \\
Gras::N & grass \\
Grenze::N & border \\
Griff::N & handle \\
Grube::N & pit \\
Grund::N & reason \\
Gr{\"o}{\ss}e::N & size \\
G{\"u}rtel::N & belt \\
Haar::N & hair \\
Haken::N & hook \\
Hals::N & neck \\
Hand::N & hand \\
Handfl{\"a}che::N & palm \\
Handtuch::N & towel \\
Haufen::N & heap \\
Haus::N & house \\
Haut::N & skin \\
Heim::N & home \\
Hemd::N & shirt \\
Herz::N & heart \\
Heu::N & hay \\
Hilfe::N & help \\
Himmel::N & sky \\
Hitze::N & heat \\
Holz::N & wood \\
Honig::N & honey \\
Horn::N & horn \\
Hose::N & trousers \\
Hunger::N & hunger \\
H{\"a}lfte::N & half \\
H{\"o}he::N & height \\
H{\"o}hle::N & cave \\
H{\"u}gel::N & hill \\
Insel::N & island \\
Jahr::N & year \\
Kamm::N & comb \\
Kampf::N & fight \\
Kante::N & edge \\
Kehle::N & throat \\
Kessel::N & kettle \\
Kiefer{[}Anatomie{]}::N & jaw \\
Kiefer{[}Baum{]}::N & pine \\
Kinn::N & chin \\
Kirche::N & church \\
Kissen::N & pillow \\
Kiste::N & box \\
Klaue::N & claw \\
Kleidung::N & clothes \\
Knie::N & knee \\
Knochen::N & bone \\
Knopf::N & button \\
Knoten::N & knot \\
Kohle::N & coal \\
Kopf::N & head \\
Korn::N & grain \\
Kraft::N & force \\
Kragen::N & collar \\
Kralle::N & claw \\
Krankheit::N & illness \\
Kreis::N & circle \\
Kreuz::N & cross \\
Krieg::N & war \\
Kummer::N & grief \\
K{\"a}lte::N & chill \\
K{\"o}rper::N & body \\
K{\"u}ste::N & coast \\
Laden::N & shop \\
Lagerfeuer::N & campfire \\
Land::N & country \\
Last::N & load \\
Laut::N & sound \\
Leben::N & life \\
Leber::N & liver \\
Leder::N & leather \\
Lehm::N & clay \\
Leine::N & leash \\
Leiter::N & ladder \\
Leute::N & people \\
Licht::N & light \\
Lied::N & song \\
Linie::N & line \\
Lippe::N & lip \\
Loch::N & hole \\
Luft::N & air \\
Lust::N & desire \\
L{\"a}nge::N & length \\
L{\"a}rm::N & noise \\
L{\"o}ffel::N & spoon \\
L{\"u}ge::N & lie \\
Macht::N & power \\
Magen::N & stomach \\
Meer::N & sea \\
Menge::N & amount \\
Messer::N & knife \\
Milch::N & milk \\
Mittag::N & noon \\
Mitte::N & middle \\
Monat::N & month \\
Mond::N & moon \\
Moor::N & moor \\
Morgen::N & morning \\
Mund::N & mouth \\
Muster::N & pattern \\
M{\"a}rchen::N & fairy tale \\
M{\"u}tze::N & cap \\
Nabel::N & navel \\
Nachricht::N & message \\
Nacht::N & night \\
Nadel::N & needle \\
Nagel::N & nail \\
Nagel{[}Anatomie{]}::N & nail \\
Nahrung::N & food \\
Name::N & name \\
Nase::N & nose \\
Nebel::N & fog \\
Nest::N & nest \\
Netz::N & net \\
Neuigkeit::N & news \\
Norden::N & north \\
Nutzen::N & profit \\
Oberschenkel::N & thigh \\
Ofen::N & stove \\
Ohr::N & ear \\
Ort::N & place \\
Osten::N & east \\
Pfad::N & path \\
Pfeil::N & arrow \\
Pilz::N & mushroom \\
Platte::N & slab \\
Platz::N & space \\
Preis::N & price \\
Puppe::N & doll \\
Quelle::N & source \\
Rand::N & fringe \\
Rauch::N & smoke \\
Raureif::N & hoarfrost \\
Rede::N & speech \\
Regal::N & shelf \\
Regen::N & rain \\
Regenbogen::N & rainbow \\
Reichtum::N & wealth \\
Reihe::N & row \\
Riemen::N & strap \\
Rinde::N & bark \\
Ring::N & ring \\
Rohr::N & pipe \\
Ruder::N & oar \\
Ruf::N & call \\
Ruhe::N & calm \\
R{\"a}tsel::N & puzzle \\
R{\"u}cken::N & back, spine \\
Saat::N & seed \\
Sache::N & thing \\
Sack::N & sack \\
Salz::N & salt \\
Sand::N & sand \\
Schaden::N & damage \\
Schale::N & husk \\
Schatten::N & shadow \\
Schaufel::N & shovel \\
Schaum::N & foam \\
Scheibe::N & slice \\
Schlaf::N & sleep \\
Schlinge::N & noose \\
Schlitten::N & sleigh \\
Schloss::N & lock \\
Schluss::N & conclusion \\
Schmerz::N & pain \\
Schmutz::N & dirt \\
Schnee::N & snow \\
Schnur::N & string \\
Schnurrbart::N & moustache \\
Schritt::N & step \\
Schuh::N & shoe \\
Schuld::N & fault \\
Schulter::N & shoulder \\
Schwanz::N & tail \\
See::N & lake \\
Sehne::N & sinew \\
Seite::N & side \\
Silber::N & silver \\
Sinn::N & meaning \\
Ski::N & ski \\
Sonne::N & sun \\
Spaten::N & spade \\
Speise::N & dish \\
Spiegel::N & mirror \\
Spiel::N & game \\
Spitze::N & tip \\
Sprache::N & language \\
Spur::N & track \\
Staat::N & state \\
Stab::N & staff \\
Stadt::N & town \\
Stamm::N & trunk \\
Stange::N & pole \\
Staub::N & dust \\
Stein::N & stone \\
Stern::N & star \\
Stiefel::N & boot \\
Stimme::N & voice \\
Stirn::N & forehead \\
Stock::N & stick \\
Stoff::N & cloth \\
Stra{\ss}e::N & road \\
Strich::N & stroke \\
Str{\"o}mung::N & current \\
Stuhl::N & chair \\
St{\"a}rke::N & strength \\
St{\"u}ck::N & piece \\
St{\"u}tze::N & bracket \\
Sumpf::N & swamp \\
Suppe::N & soup \\
S{\"u}den::N & south \\
S{\"u}nde::N & sin \\
Tag::N & day \\
Tanne::N & fir \\
Tasche::N & bag \\
Tasse::N & cup \\
Tee::N & tea \\
Teil::N & part \\
Tisch::N & table \\
Tod::N & death \\
Ton::N & tone \\
Topf::N & pot \\
Tor::N & gate \\
Traum::N & dream \\
Tropfen::N & drop \\
Tr{\"a}ne::N & tear \\
Tuch::N & scarf \\
T{\"u}r::N & door \\
Ufer::N & shore \\
Ungl{\"u}ck::N & misfortune \\
Verstand::N & mind \\
Volk::N & nation \\
Wahrheit::N & truth \\
Wald::N & forest \\
Wange::N & cheek \\
Ware::N & ware \\
Wasser::N & water \\
Weg::N & way \\
Weide::N & pasture \\
Weide{[}Baum{]}::N & willow \\
Welle::N & wave \\
Welt::N & world \\
Westen::N & west \\
Wetter::N & weather \\
Wiege::N & cradle \\
Wiese::N & meadow \\
Wind::N & wind \\
Winkel::N & angle \\
Woche::N & week \\
Wolke::N & cloud \\
Wolle::N & wool \\
Wort::N & word \\
Wunde::N & wound \\
Wunsch::N & wish \\
Wurzel::N & root \\
Zahn::N & tooth \\
Zaun::N & fence \\
Zeh::N & toe \\
Zeichen::N & sign \\
Zeit::N & time \\
Zeitung::N & newspaper \\
Zunge::N & tongue \\
Zweig::N & branch \\
Zwiebel::N & onion \\
{\"A}rmel::N & sleeve \\
{\"O}l::N & oil \\
\bottomrule
\end{supertabular}

\end{document}